\pdfoutput=1

\documentclass{article}
\usepackage{ijcai19}

\usepackage{times}
\usepackage{soul}
\usepackage{url}
\usepackage[hidelinks]{hyperref}
\usepackage[utf8]{inputenc}
\usepackage[small]{caption}
\usepackage{graphicx}
\usepackage{amsmath}
\usepackage{booktabs}
\usepackage{adjustbox}
\usepackage{pgf}
\usepackage{tikz}
\usepackage{csquotes}
\usetikzlibrary{calc,quotes,arrows.meta}
\usetikzlibrary{shapes.misc,positioning,calc,graphs,arrows}
\usepackage[ruled,vlined]{algorithm2e}
\usepackage{dsfont}
\usepackage{capt-of}
\usepackage{xcolor}
\usepackage{subcaption}
\usepackage{empheq}
\usepackage{booktabs}
\def \NN {\mathcal{N}}
\def \Ber {\mathcal{B}er}

\def \Cat {\mathcal{C}\!\mathit{at}}

\def \DD{\mathcal{D}}

\def\R{\mathds{R}}
\def \E{\mathds{E}}
\def \KL{\mathrm{KL}}

\def \H{\mathrm{H}}

\def \1{\mathds{1}}

\def \veps{\varepsilon}

\def \mbw{\mathbf{w}}
\def \mby{\mathbf{y}}
\def \mbx{\mathbf{x}}

\DeclareMathOperator*{\argmax}{arg\,max}

\title{Deep Active Learning with Adaptive Acquisition}

\author{
Manuel Hau\ss mann$^1$\and
Fred Hamprecht$^1$\And
Melih Kandemir$^2$
\affiliations
$^1$HCI/IWR, Heidelberg University, Germany\\
$^2$Bosch Center for Artificial Intelligence, Renningen, Germany\\
\emails manuel.haussmann@iwr.uni-heidelberg.de
}

\begin{document}

\maketitle

\begin{abstract}
  Model selection is treated as a standard performance boosting step in many machine learning applications. Once all other properties of a learning problem are fixed, the model is selected by grid search on a held-out validation set. This is strictly inapplicable to active learning. Within the standardized workflow, the acquisition function is chosen among available heuristics a priori, and its success is observed only after the labeling budget is already exhausted. More importantly, none of the earlier studies report a unique consistently successful acquisition heuristic to the extent to stand out as the unique best choice. We present a method to break this vicious circle by defining the acquisition function as a learning predictor and training it by reinforcement feedback collected from each labeling round. As active learning is a scarce data regime, we bootstrap from a well-known heuristic that filters the bulk of data points on which all heuristics would agree, and learn a policy to warp the top portion of this ranking in the most beneficial way for the character of a specific data distribution. Our system consists of a Bayesian neural net, the predictor, a bootstrap acquisition function, a probabilistic state definition, and another Bayesian policy network that can effectively incorporate this input distribution. We observe on three benchmark data sets that our method always manages to either invent a new superior acquisition function or to adapt itself to the a priori unknown best performing heuristic for each specific data set.
\end{abstract}

\section{Introduction} \label{sec:intro}

\begin{figure*}
  \centering
  \begin{subfigure}{\columnwidth}
    \tikzset{roundrect/.style={
        rectangle, 
        rounded corners,
        minimum size=6mm,
        draw=black,
    }}
    \centering
    \begin{tikzpicture}[thick,auto, >=Stealth, node distance=2cm,scale=0.95, every node/.style={scale=0.95}]
    \node[roundrect, align=center] (oracle) at (0,0) {\textbf{Oracle}};
    \node[roundrect, below right of=oracle, xshift=1.2cm, align=center] (classifier) {\textbf{Predictor} \\ {\footnotesize BNN}};
    \node[roundrect, below left of=oracle, xshift=-1.2cm, align=center] (guide) {\textbf{Guide}\\{\footnotesize Acquisition Score}\\ {\footnotesize (heuristic)}};
    \coordinate[above=of classifier, xshift=-0.9cm,yshift=-2cm] (corner);  
    \coordinate[above=of classifier, xshift=-0.9cm,yshift=-2cm] (corner);
    
    \draw [->] (oracle) edge [out=0, "{\footnotesize label}"] (classifier)
    (classifier) edge [out=190,in=350, "{\footnotesize prediction uncertainty}"] (guide)
    (guide) edge [in=180, near start, "{\footnotesize chosen sample}"] (oracle);
    \end{tikzpicture}
    
    \caption{Active learning pipeline.}
  \end{subfigure}
  \begin{subfigure}{\columnwidth}
    \centering
    \tikzset{roundrect/.style={
        rectangle, 
        rounded corners,
        minimum size=6mm,
        draw=black,
    }}
    \begin{tikzpicture}[thick,auto, >=Stealth, node distance=2cm,scale=0.85, every node/.style={scale=0.85}]
    \node[roundrect, align=center] (oracle) at (0,0) {\textbf{Oracle}};
    \node[roundrect, black!30!red, below left of=oracle, xshift=-0.5cm, align=center] (guide) {\textbf{Guide}\\ {\footnotesize Policy BNN}};
    \node[roundrect, black!30!red, below of=guide, align=center, font=\footnotesize] (state) {Bulk filter\\ (heuristic)};
    \node[roundrect, right of=state, xshift=2cm, align=center] (predict) {\textbf{Predictor}\\ \footnotesize BNN};
    
    \draw [->] (oracle) edge [out=0, in=90, "{\footnotesize label}"] (predict)
    (predict) edge [out=200, in=340, "{\footnotesize prediction uncertainty}"] (state)
    (state) edge [out=100, black!30!red, in=260] node[align=center, black!30!red, swap, font=\scriptsize] {probabilistic\\ environment\\ state} (guide)
    (guide) edge [out=90, in=180] node[align=center, pos=0.5, swap, font=\scriptsize] {chosen\\ sample} (oracle)
    (oracle) edge [out=340, black!30!red, in=0] node[align=center, black!30!red, pos=0.6, font=\footnotesize] {reward} (guide);
    
    \end{tikzpicture}
    \caption{Our method.}
  \end{subfigure}
  \caption{The proposed pipeline. The standard active learning pipeline is summarized as the interplay between three parts. \emph{(a)} An \emph{oracle} provides a set of labeled data for a \emph{predictor} (here a BNN) to learn on. It in turn provides predictive uncertainties to the \emph{guide}, a usually fixed, hard-coded acquisition function, which in turn communicates to the oracle which points to label next, restarting the cycle. 
  \emph{(b)} We replace the fixed acquisition function with a policy BNN that learns with a probabilistic state and reinforcement feedback from the oracle how to optimally choose the next points (new parts in red). It is thus able to adapt itself flexibly to the data set at hand.}\label{fig:figureone}
\end{figure*}
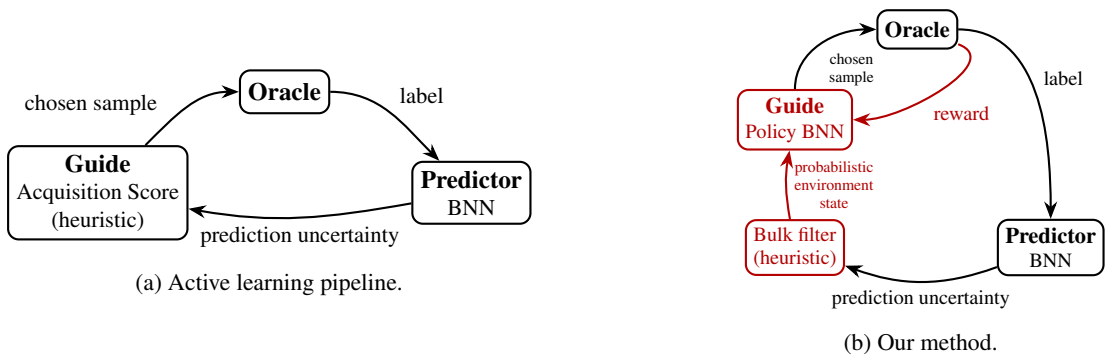

The active learning setup consists of a base predictor that chooses the order in which the data points are supposed to be labeled using an acquisition function. Contrary to the {\it tabula rasa} ansatz of the present deep learning age, the state of the art in active learning has maintained to use hand-designed acquisition functions to rank the unlabeled sample space. Different studies observed different acquisition functions to perform optimally for specific applications after evaluating various choices. The critical fact is that active learning is meant to address applications where data labeling is extremely costly and it is not possible to know the ideal acquisition function for a given application a priori. Once an acquisition function is chosen and active learning has been performed based on it, the labeling budget is already exhausted, leaving no possibility for another try with an alternative acquisition function. This limitation can only be circumvented by adapting the acquisition function to data during the active learning process, getting feedback from the impact of the previous labeling rounds on model fit. For real-world scenarios to which active learning is applicable, learning also the acquisition function is not only an option driven solely by practical concerns such as avoidance of handcrafting effort, but also an absolute necessity stemming from epistemic limitations.

The acquisition functions in active learning are surrogate models that map a data point to a value that encodes the expected contribution of observing its label to model fit. The founding assumption of the active learning setup is that evaluating the acquisition score of a data point is substantially cheaper than acquiring its ground-truth label. Hence, the acquisition functions are expected to be both computationally cheap and maximally accurate in detecting most information-rich regions of the sample space. These competing goals are typically addressed by information-theoretic heuristics. 
Possibly the most frequently used acquisition heuristic is {\it Maximum Entropy Sampling}, which assigns the highest score to the data point for which the predictor reports highest entropy (i.e.~uncertainty). This criterion builds on the assumption that the most valuable data point is the one the model is maximally unfamiliar about. While being maximally intuitive, this method remains agnostic to exploiting knowledge from the current model fit about how much the new label can impact the model uncertainty. Another heuristic with comparable reception, {\it Bayesian Active Learning by Disagreement (BALD)}~\cite{houlsby2012collaborative}, benefits from this additional information by maximizing the mutual information between the predictor output and the model parameters.
A second major vein of research approaches the active learning problem from a geometric instead of an uncertainty based perspective, e.g.~via selection of a core-set~\cite{sener2017active}.

None of the aforementioned heuristics has a theoretical superiority that is sufficient to rule out all other options. Maxent strives only to access unexplored data regions. BALD performs the same by also taking into account the expected effect of the newly explored label on the uncertainty of the model parameters. While some studies argue in favor of BALD due to this additional information it enjoys~\cite{srinivas2012information}, others prefer to avoid this noisy estimate drawn from an under-trained model~\cite{qiu2017maximum}. 
This paper presents a data-driven method that alleviates the consequences of the unsolved acquisition function selection problem. As prediction uncertainty is an essential input to acquisition heuristics, we choose a deep Bayesian Neural Net (BNN) as our base predictor. In order to acquire high-quality estimates of prediction uncertainty with an acceptable computational cost, we devise a deterministic approximation scheme that can both effectively train a deep BNN and calculate its posterior predictive density following a chain of closed-form operations. Next, we incorporate all the probabilistic information provided by the BNN predictions into a novel state design, which brings about another full-scale probability distribution. This distribution is then fed into a probabilistic policy network, which is trained by reinforcement feedback collected from every labeling round in order to inform the system about the success of its current acquisition function. This feedback fine-tunes the acquisition function, bringing about improved performance in the subsequent labeling rounds.
Figure~\ref{fig:figureone} depicts the workflow of our method.

We evaluate our method on three benchmark vision data sets from different domains and complexities: MNIST for images of handwritten digits, FashionMNIST for greyscale images of clothes, and CIFAR-10 for colored natural images. We observe in our experiments that the policy net is capable of inventing an acquisition function that outperforms all the handcrafted alternatives if the data distribution permits. In the rest of the cases, the policy net converges to the best-performing handcrafted choice, which varies across data sets and is unknown prior to the active learning experiment.

\section{The Model}\label{sec:model}
Our method consists of two major components: a predictor and a policy net guiding the predictor by learning a data set specific acquisition function. 
As the predictor, described in Section~\ref{ssec:bnn} we use a BNN, whose posterior predictive density we use to distill the system state. The policy net, another BNN, takes this state as input to decide which data points to request labels for next.\footnote{ For illustrative purposes we rely on a Central Limit Theorem approach to efficiently marginalize the weights of these BNNs. In general any approach to training a BNN which provides trustworthy predictive uncertainties could be used.} We describe this second part of the pipeline in Section~\ref{ssec:agent}. Since we introduce a reinforcement learning based method for active learning, we refer to it as \emph{Reinforced Active Learning (RAL)}. 

\subsection{Predictor: Bayesian Neural Network}\label{ssec:bnn}
Let $\DD = \{(\mbx_n, \mby_n)_{n=1}^N\}$ be a data set consisting of $N$ tuples of feature vectors $\mbx_n \in \R^m$ and labels $\mby_n \in \{0,1\}^C$ for a $C$ dimensional binary output label. Parameterizing an arbitrary neural network $f(\cdot)$ with parameters $\mbw$ following some prior $p(\mbw)$, we assume the following probabilistic model
\begin{equation}
\mbw \sim p(\mbw),\quad\mby|\mbx, \mbw \sim \prod_c^C\Ber\big (y_c|\Phi(f_c(\mbx;\mbw)\big),
\end{equation}
where $f_c$ is the $c$th output channel of the net, $\Phi(u)=\int_{-\infty}^u \NN(x|0,1) dx$, and $\Ber(\cdot|\cdot)$ is a Bernoulli distribution.
The calculation of the posterior predictive%
\begin{equation}
p({\bf y}^* | {\bf x}^*, \mathcal{D}) = \int p({\bf y}^* | {\bf x}^*, \mbw) p(\mbw  | \mathcal{D}) d\mbw
\end{equation}
involves the calculation of the posterior distribution on the latent variables, which can be accomplished by Bayes rule
\begin{equation}
p(\mbw | \DD ) = \dfrac{p({\bf Y}|{\bf X},\mbw) p(\mbw)}{\int p({\bf Y}|{\bf X},\mbw) p(\mbw) d\mbw},
\end{equation}
for $\mathbf{X} = \{\mbx_1,...,\mbx_N\}$ and $\mathbf{Y} = \{\mby_1,...,\mby_N\}$. 
As this is intractable in general we require approximate inference techniques. We aim for 
high-quality prediction uncertainties. A sampling based approach is not practical for vision-scale applications where neural nets with a large parameter count are being used. Instead we use variational inference (VI). In order to keep the calculations maximally tractable while benefiting from stochasticity $\mbw$, we formulate a normal mean-field variational posterior (which could be generalized):
\begin{equation}
q_\theta(\mbw) =  \prod_i \NN(w_i|\mu_i, \sigma_i^2), \label{eq:var_dist}
\end{equation}
where the tuple $({\mu}_{i},{\sigma}_{i}^2)$ represents the variational parameter set for weight $w_i$ of the network $f(\cdot)$ and ${\theta = \{(\mu_i, \sigma_i^2)_i\}}$. VI approximates the true intractable posterior by optimizing $\theta$ to minimize the Kullback-Leibler (KL) divergence between $q_\theta(\mbw)$ and $p(\mbw | {\bf X}, {\bf Y} )$, which boils down to minimizing the negative evidence lower bound
\begin{align}
\mathcal{L}_\mathrm{class}(\theta; \mathcal{D}) &= -\sum_{n=1}^N \E_{q_\theta(\mbw)}\big[ \log p({\bf y}_n | f({\bf x}_n; \mbw) )\big]\nonumber\\
&\qquad\qquad + \KL\big(q_\theta(\mbw)||p(\mbw)\big). 
\end{align}
In this formula, the first term on the r.h.s.~maximizes the data fit (i.e.~minimizes the reconstruction loss), and the second term penalizes divergence of the posterior from the prior, inducing the Occam's razor principle to the preferred solution. 

The modeler has control on the model families of both $q_\theta(\mbw)$ and $p(\mbw)$. Hence, choosing the prior $p(\mbw)$ suitably to the normally distributed $q_\theta(\mbw)$ assures an analytically tractable solution for the $\KL(\cdot||\cdot)$ term.\footnote{We use $p(w_i) = \NN(w_i|0, \alpha^{-1})$ with a fixed precision $\alpha$.} However, the data fit term involves a nonlinear neural net, which introduces difficulties for keeping the calculation of the expectations tractable. A major issue is that we need to differentiate this term with respect to the variational parameters $\theta$, which appear in the density $q_\theta(\mbw)$ with respect to which the expectation is taken. This problem is overcome by the reparameterization trick~\cite{kingma2013auto}, which re-formulates $q_\theta(\mbw)$ as a sampling step from a parameter-free distribution and a deterministic variable transformation. 
\begin{align}
\mathcal{L}_\mathrm{class}(\theta; \mathcal{D}) &= -\sum_{n=1}^N \E_{p(\veps)}\big[ \log p({\mby}_n | f(\mbx_n; \mbw=\mu + \sigma \veps))\big]\nonumber\\
&\qquad\qquad+ \KL\big(q(\mbw)||p(\mbw)\big),
\end{align}
where the variational parameters $\theta$ now appear only inside the expectation term, and we could take the gradient of the loss with respect to them and approximate integral in the expectation by Monte Carlo sampling. Although this will provide an unbiased estimator of the exact gradient, due to distorting the global variables of a highly-parameterized system, this estimator will have prohibitively high variance. The remedy is to postpone sampling one step further. 

Let the pre-activation and feature map of $j$th neuron of layer $l$ for data point $n$ be $b_{njl}$ and $h_{njl}$, respectively. We then have
\begin{equation}
w_{ijl} \sim \mathcal{N}(w_{ijl}|\mu_{ijl},{\sigma}_{ijl}^2),\quad b_{njl} = \sum_{i=1}^{I_{l-1}} w_{ijl} h_{n i {l-1} },
\end{equation}

as a repeating operation at every layer transition within a BNN.\footnote{The same line of reasoning directly applies to convolutional layers where the sum on $b_{njl}$ is performed in a sliding window fashion.} As $h_{n i {l-1} }$ is the sampling output of layer $l-1$, it is a deterministic input to layer $l$. Consequently, $b_{njl}$ is a weighted linear sum of $I_{l-1}$ independent normal random variables, which is another normal random variable with
\begin{equation}
b_{njl} \sim \mathcal{N} \Bigg (b_{njl} \Bigg | \sum_{i=1}^{I_{l-1}} \mu_{ijl} h_{n i {l-1} }, \sum_{i=1}^{I_{l-1}} \sigma_{ijl}^2 h_{n i {l-1} }^2   \Bigg ).
\end{equation}
We now take separate samples for local variables, further reducing the estimator variance stemming from the Monte Carlo integration. This is known as Variational Dropout~\cite{kingma2015variational}, as the process performed for the expected log-likelihood term is equivalent to Gaussian dropout with rate ${\sigma}_{ijl}^2/\mu_{ijl}^2$ for weight $w_{ijl}$.

\subsubsection{Fast Dropout and the CLT Trick}
Fast Dropout~\cite{wang2013fast} has been introduced as a technique to perform Gaussian dropout without taking samples. The technique builds on the observation that $b_{njl}$ is essentially a random variable that consists of a sheer sum of a provisionally large number of other random variables. This is a direct call to the Central Limit Theorem (CLT) that transforms the eventual distribution into a normal density, which can be trivially modeled by matching the first two moments
  \begin{align*}
  p(b_{njl}) &\approx \mathcal{N}(b_{njl} | \phi_{njl}, \lambda_{njl}^2),\quad\text{where}\\
  \phi_{njl} &= \E \Bigg[ \sum_{i=1}^{I_{l-1}} w_{ijl} h_{n i {l-1} } \Bigg ]= \sum_{i=1}^{I_{l-1}} \E[w_{ijl}] \E[ h_{ni{l-1}} ],\\
  \lambda_{njl}^2 &= \mathrm{var} \Bigg[ \sum_{i=1}^{I_{l-1}} w_{ijl} h_{n i {l-1} } \Bigg ] \\
  &= \sum_{i=1}^{I_{l-1}}  \mathrm{var}[h_{n i {l-1} }] \E[w_{ijl}]^2  + \mathrm{var}[w_{ijl}] \E[ h_{n i {l-1} }^2 ]. %
  \end{align*}
Here, $\E[w_{ijl}]=\mu_{ijl}$ and $\mathrm{var}[w_{ijl}]=\sigma_{ijl}^2$, as determined in Equation~\ref{eq:var_dist}. We also require the first two moments over of the $h_{nil-1}=r(b_{nil-1})$, for which it suffices to solve
\begin{align*}
\E[h_{n i {l-1}}] &= \int  r(b_{n i {l-1} }) p(b_{n i {l-1} }) d b_{n i {l-1} }, \\
\E[h_{n i {l-1}}^2] &= \int  r(b_{n i {l-1} })^2 p(b_{n i {l-1} }) d b_{n i {l-1} }.
\end{align*}
These two are analytically tractable for standard choices of activation functions, such as when $r(\cdot)$ is the ReLU activation and $p(b_{n i {l-1} })$ is a normal distribution~\cite{frey1999variational}. Note that
$b_{n i {l-1} }$ is either the linear activation of the input layer, a weighted sum of normals, hence another normal, or a hidden layer, which
will then similarly follow CLT and therefore already be approximated as a normal. Hence, the above pattern repeats throughout the entire network, allowing a tight closed-form approximation of the analytical solution. Below, we refer to this method as the {\it CLT trick}. 

\subsubsection{Closed-Form Uncertainty Estimation with BNNs}
Fast Dropout uses the aforementioned CLT trick only for implementing dropout. Here we extend the same method to perform variational inference by minimizing a deterministic loss, i.e.~avoiding Monte Carlo sampling altogether. Even though the CLT trick has previously been used mainly for expectation propagation, its direct application to variational inference has not been investigated prior to our work. Furthermore, the state of the art in deep active learning relies on test-time dropout~\cite{gal17a}, which is computationally prohibitive. Speeding up this process requires parallel computing on the {\it end-product}, hence reflects additional costs on the user of the model not addressable at the production time. A thus far overlooked aspect of the CLT trick is that it also allows closed-form calculation of the posterior predictive density. Once training is over, we get a factorized surrogate for our posterior. Plugging this surrogate into the predictive density formula, for a new observation ${\bf x}^*$ we get
\begin{align}
p(&y_c^*|\mbx^*, \mathcal{D}) \approx \int \Ber\big(y_c^* | \Phi(f_c(\mbx^*; \mbw) ) \big) q_\theta(\mbw) d\mbw\nonumber\\
&\approx \int \Ber\big(y_c^* | \Phi(f_c^*) \big) \NN\Big(f_c^*|g^L_c(\mbx^*),h^L_c(\mbx^*)\Big) d f_{c}^*\nonumber\\
&= \Ber\Bigg (y_c^* \Bigg | \Phi \Bigg( \dfrac{g^L_c(\mbx^*)}{ \sqrt{h^L_c(\mbx^*)+1}} \Bigg ) \Bigg),\label{eq:pred}
\end{align}
where the functions $g^L_c(\mbx^*)$ and $h^L_c(\mbx^*)$ encode the cumulative map from the input layer to the moments of the top-most layer after repetitive application of the CLT trick across the layers.\footnote{Once $g_c^L(\mbx^*)$ and $h_c^L(\mbx^*)$ are computed one could also choose a categorical likelihood and approximate the integral via sampling.} With $p(y^*_c|\mbx^*, \DD)$ is tightly approximated by an analytical calculation of a known distributional form, its high-order moments are readily available for downstream tasks, being active learning in our case.

\subsection{Guide: The Policy Net}\label{ssec:agent}

As opposed to the standard active learning pipeline, our method is capable of adapting its acquisition scheme to the characteristics of individual data distributions. Differently from earlier work on data-driven label acquisition, our method can perform the adaptation {\it on the fly}, i.e.~while the active learning labeling rounds take place. This adaptiveness is achieved within a reinforcement learning framework, where a policy net is trained by rewards observed from the environment. 
We denote the collection of unlabeled points by $\DD_u$ and the labeled ones by $\DD_l$. The variables $N_u$ and $N_l$ denote the number of data points in each respective case.

\paragraph{State.} In active learning, the label acquisition process takes place on the entire unlabeled sample set. However, a feasible reinforcement learning setup necessitates a condensed state representation. To this end, we first rank the unlabeled sample set by an information-theoretic heuristic, namely the maximum entropy criterion. As such heuristics assign similar scores to samples with similar character, consecutive samples in ranking inevitably have high correlation. In order to break the trend and enhance diversity, we follow the ranking from the top and pick up every $K$th sample until we collect $M$ samples $\{ {\bf x}_1, \cdots, {\bf x}_M \}$. We adopt the related term from the Markov Chain Monte Carlo sampling literature and refer to this process as {\it thinning}. Feeding these samples into our predictor BNN (Equation~\ref{eq:pred}), we attain a posterior predictive density estimate for each and distill the state of the unlabeled sample space in the following distributional form:
\begin{equation}
S \sim \prod_{c=1}^C \prod_{m=1}^M \mathcal{N}\left (g_c^L(\mbx_m), h_c^L(\mbx_m)\right ),
\end{equation}
where $g_c^L(\cdot)$ and $h_c^L(\cdot)$ are mean and variance of the activation an output neuron, calculated as in Equation~\ref{eq:pred}.

\paragraph{Action.} At each labeling round, a number of data points are sampled from the set $\{\mbx_{i_1},...,\mbx_{i_M}\}$ according to the probability masses assigned on them by the present policy.

\paragraph{Reward.} The straight-forward reward would be the performance of the updated predictor on a separate validation set. This, however, clashes with the constraint imposed on us by the active learning scenario. The motivating assumption is that labels are valuable and scarce, so it is not feasible to construct a separate labeled validation set large enough to get a good guess of the desired test set performance for the policy net to calculate rewards. In our preliminary experiments, we have observed that merging the validation set with the existing training set and performing active learning on the remaining sample set consistently provides a steeper learning curve than keeping a validation set for reward calculation. Hence, we abandon this option altogether. Instead, we propose a novel reward signal 
\begin{equation}
  R = R_\text{improv} + R_\text{div},
\end{equation}
 consisting of the two components detailed below.
The first component $R_\text{improv}$ assesses the improvement in data fit of the chosen point once it has been labeled. From a Bayesian perspective, a principled measure of model fit is the marginal likelihood. For a newly labeled pair $(\mbx,\mby)$ the reward is
\begin{align}
R_\text{improv} &= \prod_{c=1}^C\int \Ber\big (y_c|\Phi(f_c(\mbx; \mbw))\big ) q_\text{new}(\mbw) d\theta\nonumber\\
&\quad-\prod_{c=1}^C\int \Ber\big(y_c|\Phi(f_c(\mbx; \mbw))\big) q_\text{old}(\mbw) d\theta,
\end{align}
where $q_\text{old}(\cdot), q_\text{new}(\cdot)$ are our respective variational posteriors before and after training with the new point. The second component, $R_\text{div}$, encourages diversity across the chosen labels throughout the whole labeling round:
\begin{equation}
R_\text{div} = \frac{\# \text{unique labels requested}}{\# \text{label requests in this episode}}.
\end{equation}

\paragraph{Policy net.}
The policy net $\pi(\cdot)$ is a second BNN parameterized by $\phi\sim p(\phi)$. Compared to the classifier, taking deterministic data points as input, the policy net takes the state $S$, which follows a $C\cdot M$-dimensional normal distribution. Inputing such a stochastic input into our deterministic inference scheme is straightforward by using the first two moments of the state during the first moment-matching round. The output of the policy net, in turn, parameterizes an $M$ dimensional categorical distribution over possible actions. In order to benefit fully from the BNN character of the policy and to marginalize over the $\phi$ we again follow the approach we use for the classifier propagating the moments and first compute $M$ binary probabilities for taking action $\tilde{a}_m$ at time point $t$
\begin{equation}
p(\tilde{a}_m^t) = \E_{q(\phi)}\left[\Ber\big(\tilde{a}_m^t|\Phi(\pi_m(S_t;\phi))\big)\right],
\end{equation}
and finally normalize them to choose the action $A_t$ via
\begin{equation*}
A_t \sim \Cat(\mathbf{a}^t), \qquad\text{where~~} a^t_m = \tilde{a}^t_m\Big/{\textstyle \sum_j} \tilde{a}^t_j,
\end{equation*}
and $\Cat(\cdot)$ is a Categorical distribution.

\paragraph{Algorithm.} We adopt the episodic version of the standard REINFORCE algorithm~\cite{williams1992simple} to train our policy net. We use a moving average over all the past rewards as the baseline. A labeling episode consists of choosing a sequence of points to be labeled (with a discount factor of $\gamma=0.95$) after which the BNN is retrained and the policy net takes one update step. We parameterize the policy $\pi_\phi(\cdot|S_t)$ itself by a neural network with parameters $\phi$. Our method iterates between labeling episodes, training the policy net $\pi$, and training the BNN $f$ until the labeling budget is exhausted. The pseudocode of our method is given in Algorithm~\ref{alg:pseudo}. 

\begin{algorithm}[tb]
  \SetAlgoLined	
  \KwIn{$\DD = \{\DD_u, \DD_l\}$, labeling budget, state size~$M$, policy $\pi_\phi$, net $f_\theta$, episode length $T$}
  \BlankLine
  \tcp{Train an initial net}
  train $f_\theta$ on $\DD_l$ as described in Section~\ref{ssec:bnn}\\
  \While{budget available}{
    \tcp{The labeling episode}
    generate state distribution $S_0$ from $\DD_u$\\
    \For{$t\in {1,...,T}$}{
      sample $A_t$ via  $\pi_\phi(S_{t-1})$\\
      $\DD_l \leftarrow \DD_l \cup\{\text{data point selected via $A_t$}\}$\\
      $\DD_u \leftarrow \DD_u \backslash\{\text{data point selected via $A_t$}\}$\\
      generate state distribution $S_t$ from $\DD_u$
    }
    \tcp{Update the agent and net}
    train $f_\theta$ on $\DD_l$\\
    compute rewards $R_\text{div}, R_\text{improv}$ and returns $G_t$\\
    update $\phi$ via gradient descent on $G_t\nabla_\phi \left(\log\pi_\phi (A_t|S_t) + \KL\big(q(\phi)||p(\phi)\big)\right)$
  }  		
  \caption{The RAL training procedure}\label{alg:pseudo}
\end{algorithm}

\section{Experiments}\label{sec:experiments}

As RAL is the first method to adapt its acquisition function while active learning takes place, its natural reference model is the standard active learning setup with a fixed acquisition heuristic.\footnote{see \url{github.com/manuelhaussmann/ral} for a reference pytorch implementation of the proposed model.} 
We choose the most established two information-theoretic heuristics: Maximum Entropy Sampling (Maxent) and BALD. Gal~\textit{et.al.}~\shortcite{gal17a} already demonstrated how BNNs (in their case with fixed Bernoulli dropout) provide an improved signal to acquisition functions that can be used to improve upon using predictive uncertainty from deterministic nets.
We will use our own BNN formulation for both RAL as well as these baseline acquisition functions, to give them access to the same closed-form predictive uncertainty and to to ensure maximal comparability between our model and the baselines by having an absolutely identical architecture and training procedure for all methods.
For Maxent one selects the point that maximizes the predictive entropy,
\begin{align*}
\argmax_{(x,y) \in \DD_u}~&\H[p(y|x, \DD_l)], \qquad\text{where}\\
\H[p(y|x, \DD_l)]&=-\sum_{c=1}^C p(y=c|x,\DD_l)\log p(y=c|x, \DD_l),
\end{align*}
while  BALD chooses the point that maximizes the expected reduction in posterior entropy, or equivalently
\begin{align*}
\argmax_{(x,y) \in \DD_u} H[p(y|x,\DD_l)] - \E_{p(\mbw|\DD_l)}\big[H[p(y|x, \mbw)]\big]. %
\end{align*}
We can compute maximum entropy as well as the first of the two BALD terms in closed form, while we calculate the second term of BALD via a sampling based approach.
We also include random sampling as a---on our kind of data rather competitive---baseline and evaluate on three data sets to show the adaptability of the method to the problem at hand. 

\paragraph{Experimental Details.}
To evaluate the performance of the proposed pipeline,
we take as the predictor is a standard LeNet5 sized model (two convolutional layers of 20, 50 channels and two linear layers of 500, 10 neurons) and as the guide a policy net consisting of two layers with 500 hidden neurons.
We use three different image classification data sets to simulate varying difficulty while keeping the architectures and hyperparameters fixed. MNIST serves as a simple data set containing greyscale digits,  FashionMNIST is a more difficult data set of greyscale clothing objects, and CIFAR-10 finally is a very difficult data set given the small classifier depth that requires the classification of colored natural images. 
 The assumption of active learning that labels are scarce and expensive also entails the problem that a large separate validation set to evaluate and finetune hyperparameters is not feasible.
Both nets are optimized via Adam~\cite{kingma2014adam} using their suggested hyperparameters. 
In general we followed the assumption that an AL setting does not allow us to reserve valuable labeled data for hyper-parameter optimization so that they all remain fixed to the common defaults in the literature. 
The predictor is trained for 30 epochs between labeling rounds (labeling five points per round), while the policy net gets one update step after each round. To simulate the need to avoid a costly retraining after each labeling round the predictor net is initialized to the learned parameters from the last one, with a linearly decreasing learning rate after each round. 
In each experiment the state is constructed by ranking the unlabeled data according to their predictive entropy and then taking every twentieth point until $M=50$ points. Since all three data sets consider a ten class classification problem, the result is a $500$ dimensional normal distribution as the input to the policy network. We stop after having collected 400 points starting from an initial set of 50 data points.

\paragraph{Results.} We summarize the results in Table~\ref{tab:results}. RAL can learn to adapt itself to the data set at hand and can always outperform the baselines. Note that our central goal is to evaluate the relative performance of RAL and the baselines in these experiments and not the absolute performance. For a real world application one would use deeper architectures for more complex data sets, incorporate pretrained networks from similar labeled data sets, and use data augmentation to make maximal use from the small labeled data. Further benefits would come from using semi-supervised information, e.g.\ by assigning pseudo-labels to data points the classifier assigns a high predictive certainty~\cite{wang2017cost}. Such approaches would significantly improve the classifier performance for all models, but since they would blur the contribution of the respective acquisition function, we consciously ignore them here. 
Note that although RAL uses a thinned Maxent ranking to generate its state, it can improve upon that strategy in every case. An ablation study showed that while the thinning process can improve over the plain Maxent in some settings if one were to use it as a fixed strategy, it is not sufficient to explain the final performance difference between RAL and Maxent. 
REINFORCE owes its success to the bulk filtering step, which substantially facilitates the RL problem by filtering out a large portion of the search space. The simplified problem can thus be improved within a small number of episodes. More interactions with the environment would certainly bring further improvement at the expense of increasing labeling cost. We here present only a proof-of-concept for the idea that can improve on the feedback-free AL even within limited interaction rounds. Further algorithmic improvements are worthwhile investigating as future work, such as applying TRPO~\cite{schulman2015trust} or PPO~\cite{schulman2017proximal} in place of vanilla REINFORCE.

\begin{table}
  \centering
  \begin{adjustbox}{max width=0.95\columnwidth}
    \begin{tabular}{crrr}
      \toprule
      &     \multicolumn{1}{c}{\textsc{MNIST}}      &  \multicolumn{1}{c}{\textsc{FashionMNIST}}   &     \multicolumn{1}{c}{\textsc{CIFAR-10}}     \\ \midrule
      Random   &     $10.41 \pm 2.28$     &     $24.64 \pm 0.48$     &     $69.78 \pm 0.69$      \\
      Maxent   &     $8.61 \pm 1.25$      &     $25.72\pm 1.28$      &     $69.80 \pm 0.32$      \\
      BALD    &     $6.91 \pm 0.23$      &     $26.85\pm 0.49$      &     $69.69 \pm 1.69$      \\
      RAL (ours) & $\mathbf{6.81 \pm 0.99}$ & $\mathbf{23.69\pm 0.73}$ & $\mathbf{68.96 \pm 1.03}$ \\ \bottomrule
    \end{tabular}
  \end{adjustbox}
  \caption{{Results.} The table gives the average final error ($\pm$ one standard deviation over five runs) after labeling 400 labeled points.}\label{tab:results}  
\end{table}

\section{Related Work}\label{sec:relwork}

The gold standard in AL methods has long remained to base on hard-coded and hand-designed acquisition heuristics (see~\cite{settles2012active} for a review). 
A first extension is to not limit oneself to one heuristic, but to learn how to choose between multiple ones, e.g.~by a bandit  gorithm~\cite{baram2004online,chu2016can,hsu2015active} or a Markov Decision Process~\cite{ebert2012ralf}. However this still suffers from the problem of being limited to existing heuristics.

A further step to gain more flexibility is to formulate the problem as a meta-learning approach. The general idea~\cite{fang2017learning,konyushkova17,pang2018meta} is to use a set of labeled data sets to learn a general acquisition function that can either be applied as is to the target data set or finetuned on a sufficiently similar set of labeled data. Our approach differs from those attempts insofar as we learn the acquisition function based solely on the target data distribution while the data is labeled. 
If we take the scarcity of labels serious we can't allow ourselves the luxury of a separate large validation set to adapt a general heuristic. A separate large enough validation set also could not outperform the simple ablation study of allowing simpler acquisition functions that do not need a separate data set to instead combine that set with the labeled data they are training on. This is simply due to that as long as little labeled data is available the gain from being able to learn from  extra data tends to outweigh the benefit one would get by a complicated acquisition function, and as soon as data becomes more abundant the effectiveness of any active learning method sharply. 
We therefore discard them from comparative analysis.
A related area is the field of metareasoning~\cite{callaway2017learning}, where an agent has to learn how to request based on a limited computational budget.

Alongside the sampling-based alternatives for BNN inference, which are already abundant and standardized~\cite{blundell2015weight,gal2016dropout,kingma2015variational,louizos2017bayesian,molchanov2017variational}, deterministic inference techniques are also emerging. While direct adaptations of expectation propagation are the earliest of such methods~\cite{Gast_2018_CVPR,hernandez2015probabilistic}, they do not yet have a widespread reception due to their relative instability in training. This problem arises from the fact that EP do not provide any convergence guarantees, hence an update might either improve or deteriorate the model fit on even the training data. Contrarily, variational inference maximizes a lower bound on the log-marginal likelihood. 
Early studies exist on deterministic variational inference of BNNs~\cite{kandemir2018sampling,wu2018fixing}. However, neither quantifies the uncertainty quality by using the posterior predictive of their models for a downstream application. 
Earlier work that performs active learning with BNNs does exist~\cite{hernandez2015probabilistic,gal17a,pmlr-v80-depeweg18a}. However, all of these studies use hard-coded acquisition heuristics. 

Our state construction method that forms a normal distribution from the posterior predictives of data points shortlisted by a bootstrap acquisition criterion is novel for the active learning setting. Yet, it has strong links to model-based reinforcement learning methods that propagate uncertainties through one-step predictors along the time axis~\cite{deisenroth2011pilco}.

\section{Conclusion}
We introduce a new reinforcement based method for labeling criterion learning.  It is able to learn how to choose points in parallel to the labeling process itself, instead of requiring large already labeled subsets to learn on in an off-line setting beforehand. We achieve this by formulating the classification net, the policy net as well as the state probabilistically. We demonstrate its ability to adapt to a variety of qualitatively different data set situations performing similar to or even outperforming handcrafted heuristics. In the future work, we plan to extend the policy net with a density estimator that models the input data distribution so that it can also take the underlying geometry into account, making it less dependent on the quality of the probabilities. 

\newpage

\bibliographystyle{named}
\bibliography{ijcai19}

\end{document}